\documentclass[sigconf]{acmart}

\AtBeginDocument{%
  \providecommand\BibTeX{{%
    \normalfont B\kern-0.5em{\scshape i\kern-0.25em b}\kern-0.8em\TeX}}}

\copyrightyear{2021}
\acmYear{2021}
\setcopyright{acmcopyright}
\acmConference[MM '21]{Proceedings of the 29th ACM International Conference on Multimedia}{October 20--24, 2021}{Virtual Event, China}
\acmBooktitle{Proceedings of the 29th ACM International Conference on Multimedia (MM '21), October 20--24, 2021, Virtual Event, China}
\acmPrice{15.00}
\acmDOI{10.1145/3474085.3475317}
\acmISBN{978-1-4503-8651-7/21/10}

\usepackage{epsfig,graphicx,amsfonts,xcolor,multirow,amsmath,balance}
\usepackage{hyperref}
\usepackage{hyperxmp}
\usepackage{mycommand}

\usepackage{amsmath}
\usepackage{bm}

\begin{document}
\fancyhead{}

% \title{Exploring Sequence Feature Alignment and Bipartite Matching Consistency for Domain Adaptive Object Detection}
\title{Exploring Sequence Feature Alignment for Domain Adaptive Detection Transformers}

\finaltrue

\iffinal

\author{Wen Wang$^{1}$, Yang Cao$^{1,2}$, Jing Zhang$^{3}$, Fengxiang He$^{4}$, Zheng-Jun Zha$^{1}$}
\author{Yonggang Wen$^5$, Dacheng Tao$^4$}
\thanks{This work was done during Wen Wang's internship at JD Explore Academy. 

Corresponding Author: Yang Cao, Jing Zhang}
\affiliation{%
  \institution{
  \textsuperscript{\rm 1} University of Science and Technology of China \\
  \textsuperscript{\rm 2} Institute of Artificial Intelligence, Hefei Comprehensive National Science Center \\
  \textsuperscript{\rm 3} The University of Sydney, 
  \textsuperscript{\rm 4} JD Explore Academy, China,
  \textsuperscript{\rm 5} Nanyang Technological University
  }\country{}
%   \textsuperscript{\rm 5}Ningbo Research Institute, Zhejiang University}
}
\email{
wangen@mail.ustc.edu.cn, 
{forrest, zhazj}@ustc.edu.cn,
jing.zhang1@sydney.edu.au,
}
\email{
{hefengxiang, taodacheng}@jd.com, 
% {fengxiang.f.he, dacheng.tao}@gmail.com,
ygwen@ntu.edu.sg
}

\else

\author{Anonymous ACM Multimedia Submission}
\affiliation{\institution{Paper ID 850}}

\renewcommand{\shortauthors}{Anonymous}

\fi

\begin{abstract}
Detection transformers have recently shown promising object detection results and attracted increasing attention. However, how to develop effective domain adaptation techniques to improve its cross-domain performance remains unexplored and unclear. In this paper, we delve into this topic and empirically find that direct feature distribution alignment on the CNN backbone only brings limited improvements, as it does not guarantee domain-invariant sequence features in the transformer for prediction. To address this issue, we propose a novel Sequence Feature Alignment (SFA) method that is specially designed for the adaptation of detection transformers. Technically, SFA consists of a domain query-based feature alignment (DQFA) module and a token-wise feature alignment (TFA) module. In DQFA, a novel domain query is used to aggregate and align global context from the token sequence of both domains. DQFA reduces the domain discrepancy in global feature representations and object relations when deploying in the transformer encoder and decoder, respectively. Meanwhile, TFA aligns token features in the sequence from both domains, which reduces the domain gaps in local and instance-level feature representations in the transformer encoder and decoder, respectively. Besides, a novel bipartite matching consistency loss is proposed to enhance the feature discriminability for robust object detection. Experiments on three challenging benchmarks show that SFA outperforms state-of-the-art domain adaptive object detection methods. Code has been made available at: \url{https://github.com/encounter1997/SFA}.

\end{abstract}




\maketitle

\section{Introduction}
\label{sec:intro}

\begin{figure}[t]
  \centerline{\includegraphics[width=0.95\linewidth]{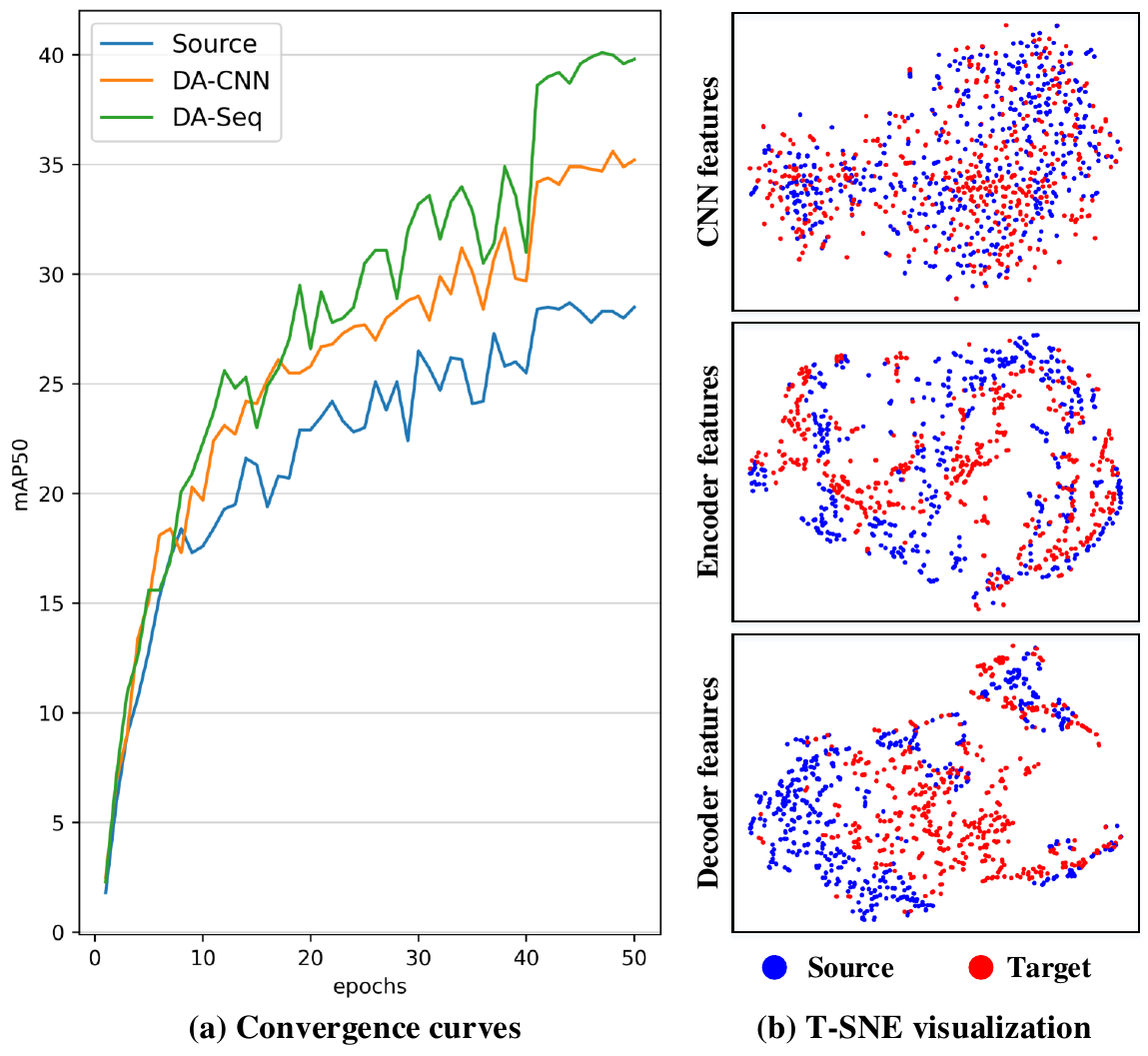}}
  \caption{
  (a) Performance comparison between feature distribution alignment applied on the CNN backbone and transformer (denoted as DA-CNN and DA-Seq, respectively). Both methods are built on the Deformable DETR~\cite{zhu2020deformable} with a ResNet-50 backbone and evaluated on the Cityscapes~\cite{cityscapes} to Foggy Cityscapes~\cite{foggy} scenario. 
%   More details are shown in the Supplementary Material. 
  (b) T-SNE visualization of features extracted by DA-CNN. While the CNN features are well-align, the sequence feature extracted by transformer encoder and decoder can be easily separated by domain.
  }
  \label{fig:figure1}
  \Description{pipeline}
  \vspace{-4mm}
\end{figure}

Object Detection is one of the fundamental tasks in computer vision and is an essential and prepositive step for multimedia applications like captioning~\cite{zha2019context, zhang2020object}, visual grounding~\cite{liu2019learning}, texture analysis~\cite{zhai2019deep, zhai2020deep}, person re-identification / person search~\cite{zheng2020exploiting, zheng2021group, zheng2020hierarchical}, \etc. 
% visual grounding, visual-language navigation, video content analysis, \etc.
Tremendous progress has been made by CNN-based object detection methods in the last decade, \eg, Faster RCNN~\cite{faster-rcnn}, SSD~\cite{ssd} and FCOS~\cite{tian2019fcos}. However, they rely on heuristics like non-maximum suppression (NMS), and are not fully end-to-end. Recently, detection transformers, \eg, DETR~\cite{detr} and Deformable DETR~\cite{zhu2020deformable}, have attracted increasing attention due to their merits of fully end-to-end pipeline and promising performance. While effective, all these methods rely heavily on the labeled training data, and suffer significant performance drops when the test data and training data are sampled from different distributions, due to domain shifts. %~\cite{}.

To tackle this problem, unsupervised domain adaptive object detection (DAOD)~\cite{dafaster} attempts to train an object detector on the labeled source domain that can be generalized to the unlabeled target domain. Existing DAOD methods~\cite{dafaster, kim2019self, hsu2020every} have achieved significant progress in improving the cross-domain performance for specific object detection models, such as based on Faster RCNN, SSD, and FCOS. With the recent surge of detection transformers, it is natural to ask, can we empower them with such a capability to perform accurate object detection in cross-domain scenarios?

A vanilla idea to solve this problem is to apply adversarial feature distribution alignment~\cite{dann} on features extracted by the CNN backbone~\cite{maf, strong-weak}. As shown in Figure~\ref{fig:figure1} (a), direct feature alignment on the CNN backbone (denoted as DA-CNN) does improve the detection transformer’s cross-domain performance, yet the improvement is limited. We argue that this is because the feature distribution alignment on the CNN backbone does not guarantee domain-invariant sequence features in the subsequent transformer, which are directly utilized for prediction. In Figure~\ref{fig:figure1} (b), we visualize the distribution of features extracted by DA-CNN. While the source and target features extracted by the CNN backbone are well-aligned, the sequence features extracted by the transformer encoder can be separated by domain. Moreover, as the network goes deeper, the distribution gaps in the decoder feature become even more significant. As a result, the detection transformer obtains inferior performance based on the shifted sequence features.

To tackle this problem, we propose a novel sequence feature alignment (SFA) method that is specially designed for the domain adaptation of detection transformers. SFA consists of a domain query-based feature alignment (DQFA) module and a token-wise feature alignment (TFA) module upon the transformer structure. Specifically, the DQFA utilizes a novel domain query to aggregate global context from the sequence for feature alignment. When applied to the transformer encoder and decoder, DQFA alleviates the domain gaps on global-level and object relations, respectively. Meanwhile, the TFA focus on feature alignment of each token in the sequence, it effectively closes domain gaps at local- and instance-level when applied to the encoder and decoder, respectively. Besides, a novel bipartite matching consistency loss is proposed to regularize the detection transformer and improve its discriminability for robust object detection. Experiments on three challenging benchmarks show that SFA significantly improves detection transformers' cross-domain performance and outperforms existing DAOD methods built on various object detectors. The main contribution of this paper can be summarized as follows:
% \vspace{-1mm}
\begin{itemize}
  \item	We dedicate to improving detection transformers' cross-domain performance, which is still unexplored and unclear. Empirically, we observe that direct feature distribution alignment on the CNN backbone only brings limited improvements, as it does not guarantee domain-invariant sequence features in the transformer for prediction.
  \item We propose Sequence Feature Alignment (SFA) that is specially designed for domain adaptation of detection transformers. It consists of a domain query-based feature alignment (DQFA) module and a token-wise feature alignment (TFA) module for aligning sequence features on a global- and local-level, respectively. Moreover, we provide technical insight on the explicit meanings of these modules when applied to the transformer encoder and decoder.
  \item A novel bipartite matching consistency loss is proposed to further regularize the sequence features and improve the discriminability of the detection transformers. 
  \item Extensive experiments on three challenging domain adaptation scenarios, including weather adaptation, synthetic to real adaptation, and scene adaptation, verify the effectiveness of our method, where SFA outperforms existing DAOD methods and achieves state-of-the-art (SOTA) performance. % (Theoretical Insights)
\end{itemize}

\section{Related work}
\label{sec:related}
\subsection{Object Detection}
% \paragraph{Object Detection}
Object detection is one of the fundamental tasks in computer vision~\cite{chen2020recursive, zhang2020empowering}. Representative deep learning-based object detectors can be roughly categorized as two-stage methods, \eg, Faster RCNN~\cite{faster-rcnn}, and single-stage methods, \eg, YOLO~\cite{yolo} and SSD~\cite{ssd}. While the former ones generally show better performance, the latter ones are faster during inference. Although significant progress has been made, these object detectors are not fully end-to-end and heavily rely on hand-crafted components, such as anchor box generation and non-maximum suppression (NMS) post-processing. Recently, DETR~\cite{detr} provides a simple and clean pipeline for object detection. It views object detection as a direct set prediction problem and explores transformer~\cite{vaswani2017attention, ding2020self, shen2021locate, shen2021trigger} and bipartite matching for effective object detection. The success of DETR brought the recent surge of detection transformers. Deformable DETR~\cite{zhu2020deformable} proposes a novel deformable attention, which speeds up model training with learnable sparse sampling and improves model performance by integrating multi-scale features. UP-DETR~\cite{up-detr} introduces a novel self-supervised pre-training scheme to improve the performance of DETR with faster convergence. 

\subsubsection{Formulation of Detection Transformers}

The detection transformer models generally consist of three parts: the CNN backbone for base feature extraction, the transformer for sequence feature modeling, and feed-forward network (FFN) for prediction. The CNN backbone extracts hierarchical feature representation ${\{f^l\}}_{l=1}^L$ from the input images, where $L$ is the number of feature levels and $f^{l} \in \mathbb{R}^{H^{l} \times W^{l} \times C^{l}}$ is the $l$-th feature map. 
% $H^{l}$, $W^{l}$ and $C^{l}$ are corresponding height, width and channel number, respectively. 
Afterwards, the hierarchical features are flatten and embedded to form an one-dimensional sequence $f_{e} \in \mathbb{R}^{N \times C}$, whose length is $N=\sum_{l=1}^{L} H^{l} W^{l}$ and the dimension of feature embeddings is $C$. The input of the transformer is denoted as ${z}_0$, which is $f_{e}$ augmented with explicit embeddings such as positional embedding and level embedding~\cite{zhu2020deformable}.

The transformer consists of an encoder and a decoder. The encoder is a stack of $L_{enc}$ encoder layers. Each encoder layer $\text {EncLayer}_{\ell}$ takes previous layer’s output ${z}_{\ell-1}$ and the sample position reference ${p}_{\ell}$ as input and outputs the encoded sequence feature ${z}_{\ell}$ as follows,
\begin{equation}
    {z}_{\ell}=\text {EncLayer}_{\ell}\left({z}_{\ell-1}, {p}_{\ell}\right), \quad \ell=1 \ldots L_{e n c}.
\end{equation}
Similarly, the decoder is a stack of $L_{dec}$ decoder layers. The input token sequence in the transformer decoder is initialized as $q_{0}$, each decoder layer $\text {DecLayer}_{\ell}$ takes the previous decoder layer’s output ${q}_{\ell-1}$, the sample position reference ${p}_{\ell}$, and the encoder output ${z}_{L_{enc}}$ as input and outputs the decoded sequence features as follows,
\begin{equation}
    {q}_{\ell}=\text{DecLayer}_{\ell}\left({q}_{\ell-1}, {p}_{\ell-1}, {z}_{L_{e n c}}\right), \quad \ell=1 \ldots L_{{dec}}.
\end{equation}

Detection transformers~\cite{detr, zhu2020deformable} usually adopt deep supervision~\cite{lee2015deeply} to facilitate model training. Classification probability vectors and bounding boxes are predicted based on the output of each decoder layer by the FFN, and are used to compute the auxiliary loss on the source domain. We denote the supervised loss on the source domain as $\mathcal{L}_{{det}}$, which is defined in~\cite{detr, zhu2020deformable}.

\subsection{Domain Adaptive Object Detection}
Domain adaptive object detection (DAOD) has been raised very recently for unconstrained scenes\cite{dafaster}. DAF~\cite{dafaster} adopts adversarial feature alignment~\cite{dann, zhang2019category} at both image-level and instance-level.
SWDA~\cite{strong-weak} adopts strong alignment on local features and weak alignment on global features from the CNN backbone. SCDA~\cite{scda} aligns region-level features across domains via grouping instances into regions. MTOR~\cite{mtor} integrates object relations into the measure of consistency cost between teacher and student modules for adaptation. Xu \etal \cite{xu2020exploring} explore image-level categorical regularization and categorical consistency regularization for robust detection across domains. GPA~\cite{gpadet} explores graph-induced prototype and class reweighted contrastive loss for effective feature alignment and adaptation. While these methods are built on the seminal Faster RCNN~\cite{faster-rcnn}, Kim \etal \cite{kim2019self} propose an SSD-based domain adaptive object detector and explore effective self-training and adversarial background regularization for adaptation. Hsu \etal \cite{hsu2020every} build an adaptive object detector base on FCOS. They estimating pixel-wise objectness and centerness and adopting center-aware feature alignment to close the domain gaps. Although significant progress has been made, most of these methods rely on specific model architecture, \eg, Faster RCNN, SSD, and FCOS, therefore cannot be directly applied to detection transformers. In this paper, we dedicate to improve detection transformers' cross-domain performance, which is still unexplored and unclear.

\begin{figure*}
  \centerline{\includegraphics[width=17cm]{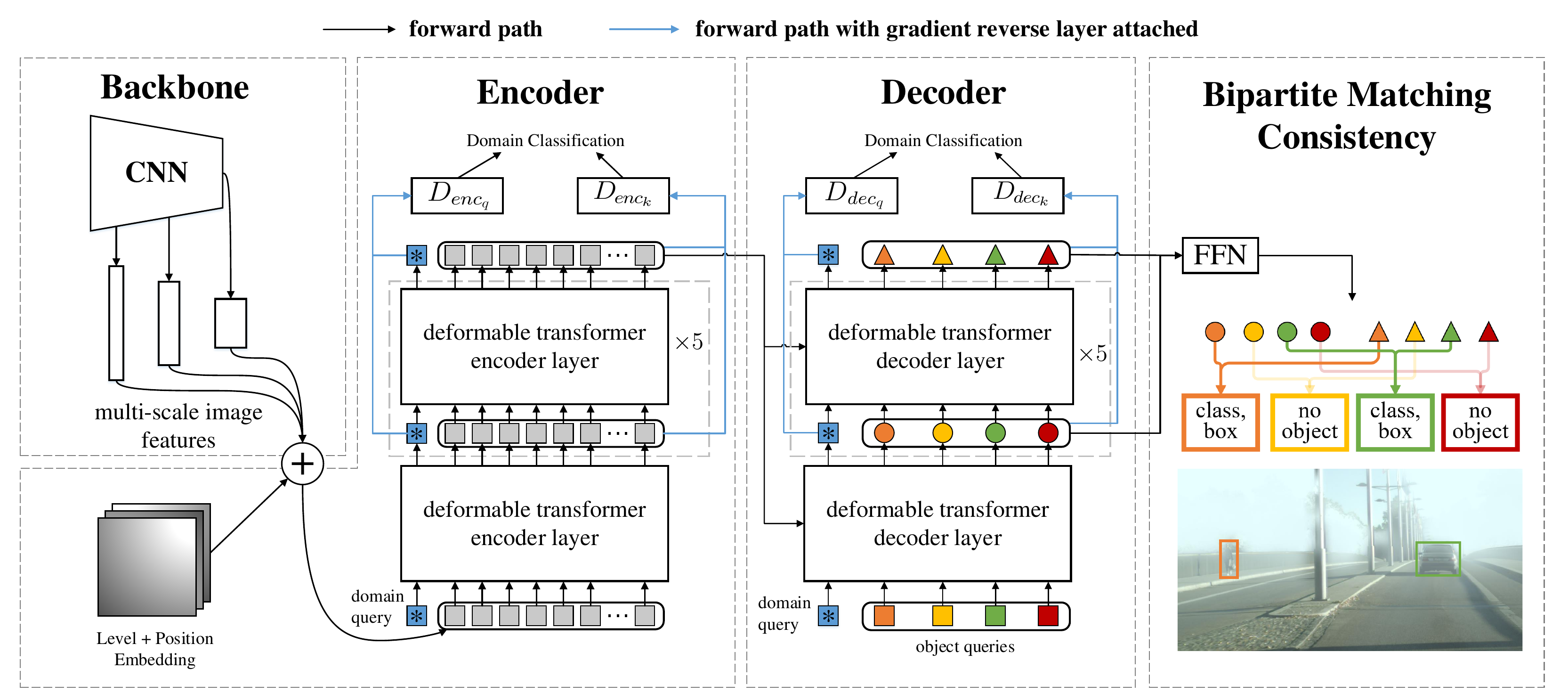}}
  \caption{Diagram of our sequence feature alignment (SFA) for domain adaptive detection transformers. Domain query-based feature alignment and token-wise feature alignment tackles global and local domain gaps, respectively. Moreover, a bipartite matching consistency loss is proposed to improve the model's discriminability. Here object queries with the same color are matched to the same ground-truth object, while object queries with the same shape come from the same decoder layer. }
  \label{fig:model}
  \Description{model}
\end{figure*}

\section{Methods}
\label{sec:methods}
This section introduces our Sequence Feature Alignment (SFA) for domain adaptive detection transformers. In unsupervised domain adaptation, the training data includes the labeled source images and the unlabeled target images. Our goal is to train an object detector on the training data that can generalize to the target domain. 

As described in Section~\ref{sec:intro}, feature distribution alignment on the CNN backbone only brings limited improvements, as it does not guarantee domain-invariant sequence features in transformer, which are directly utilized for final prediction. To solve the problem, we explore effective distribution alignment on the sequence features in transformer. Specifically, domain query-based sequence feature alignment (Section~\ref{subsec:query}) and token-wise sequence feature alignment (Section~\ref{subsec:token}) equipped with a hierarchical domain adaptation strategy (Section~\ref{subsec:proda}) are proposed. Moreover, in Section~\ref{subsec:consist}, we explore bipartite matching consistency to constrain the sequence features, which further improve detection transformer's discriminability on the target domain. The overall framework of our SFA is shown in Figure~\ref{fig:model}. We use Deformable DETR~\cite{zhu2020deformable} as the default detection framework, while our sequence feature alignment can also be applied to other detection transformer methods, \eg, DETR~\cite{detr}.

\subsection{Domain Query-based Feature Alignment}
\label{subsec:query}

To obtain domain-invariant sequence features, we first propose the domain query-based feature alignment to align source and target features from a global perspective. Specifically, on the encoder side, a domain query embedding ${q}_{{d}}^{{enc}}$ is concatenated with the token sequence to form the input ${z}_{0}$ to transformer encoder, $i.e.$,
\begin{equation}
\begin{aligned}
    {z}_{0}= \left[{q}_{{d}}^{{enc}} ; {f}_{e}^{1} ; {f}_{e}^{2} ; \cdots ; {f}_{e}^{{N}}\right]+{E}_{{pos }}+{E}_{{level}},
\end{aligned}
\end{equation}
where ${E}_{{pos }} \in \mathbb{R}^{(N+1) \times C}$ is the positional embedding, ${E}_{{level }} \in \mathbb{R}^{(N+1) \times C}$ is the feature level embedding~\cite{zhu2020deformable}.

In the encoding process, the domain query aggregates domain-specific features from the whole sequence in an adaptive manner. It encodes the global context in input images and puts more concentration on tokens with significant domain gaps. We input this domain query into a domain discriminator $D_{{enc}_{{q}}}$ for feature alignment, $i.e.$,
\begin{equation}
    \mathcal{L}_{{enc}_{q}}^{\ell}=d \log {D}_{{enc}_{{q}}}\left({z}_{\ell}^{0}\right)+(1-d) \log \left(1-{D}_{{enc }_{{q}}}\left({z}_{\ell}^{0}\right)\right), 
\end{equation}
where $\ell=1 \ldots L_{ {enc}}$ indexes layers in the encoder, $d$ is the domain label, which values 0 for source images and 1 for target images. Similarly, we concatenate a domain query ${q}_{{d}}^{{dec}}$ with the object queries to obtain the input sequence to the transformer decoder:
\begin{equation}
    {q}_{0}=\left[{q}_{{d}}^{{dec}} ; {q}^{1} ; {q}^{2} ; \cdots ; {q}^{M}\right]+{E}_{{pos}}^{'}, \quad 
\end{equation}
where ${E}_{{pos}}^{'} \in \mathbb{R}^{\left({M}+1\right) \times C}$ is the positional embedding and ${q}^{i}$ is the $i$-th object query in the sequence.
% todo: the object query and positional embedding are converse in DETR paper
In the decoding process, the domain query fuses context features from each object query in the sequence, which explicitly models the object relations. We feed the domain query to domain discriminator ${D}_{{dec}_{{k}}}$ for feature alignment:
\begin{equation}
    \mathcal{L}_{dec_{q}}^{\ell}= d \log {D}_{{dec}_{{k}} }\left({q}_{\ell}^{0}\right)+(1-d) \log \left(1-{D}_{{dec}_{{q}}}\left({q}_{\ell}^{0}\right)\right),
\end{equation}
where $\ell=1 \ldots L_{dec}$ indexes the layers in the transformer decoder.

\paragraph{Remarks.} Although both encoder and decoder adopt the domain query to perform sequence feature alignment for reducing domain gaps, it should be noted that they have different implications. Specifically, since the sequence features in the encoder are derived from feature maps extracted by the CNN backbone, the domain query aggregates global features that reflect scene layout in images for alignment. While the sequence features in the decoder represent object instances, thus the domain query encodes object relationships for adaptation. Moreover, with the help of the attention mechanism and adversarial learning, the domain query-based feature alignment adaptively puts more effort on aligning features with significant domain gaps, while less effort on features with smaller ones.

\subsection{Token-wise Feature Alignment}
\label{subsec:token}

The domain query-based global feature alignment can effectively close global domain gaps in scene layout and inter-object relationships, but it struggles to address the domain shift caused by local texture and style. To solve this problem, we propose token-wise sequence feature alignment and apply it to both encoder and decoder.

Specifically, each token embedding in the encoder sequence are fed into a domain classifier ${D}_{{enc}_{{k}}}$ for adversarial feature alignment:
% \vspace{-5mm}
\begin{equation}
    \mathcal{L}_{e n c_{k}}^{\ell}=-\frac{1}{N}\sum_{i=1}^{N}\left[d \log {D}_{{enc}_{{k}}}\left({z}_{\ell}^{i}\right)+(1-d) \log \left(1-{D}_{{enc}_{{k}}}\left({z}_{\ell}^{i}\right)\right)\right].
\end{equation}
Similarly, a domain discriminator $D_{\operatorname{dec}_{k}}$ is attached on the decoder side to align each token embedding in the decoder sequence, $i.e.$,
\begin{equation}
    \mathcal{L}_{d e c_{k} }^{\ell}=-\frac{1}{M}\sum_{i=1}^{M}\left[d \log D_{{dec}_{k}}\left({q}_{\ell}^{i}\right)+(1-d) \log \left(1-D_{{dec}_{q}}\left({q}_{\ell}^{i}\right)\right)\right].
\end{equation}
\paragraph{Remarks.}
Although both encoder and decoder adopt the token-wise sequence feature alignment, it should be noted that they have different implications. Specifically, since each token in the encoder sequence represents a local area of the image, the token-wise sequence feature alignment here alleviates domain gaps caused by local texture, appearance, \etc. By contrast, each token at the decoder side represents an individual object, therefore, the token-wise sequence feature alignment closes domain gaps at the instance level. 

It should be noted that the domain query-based feature alignment cannot be replaced by the token-wise feature alignment. Although tokens in transformer also aggregate global features to some extent, they are generated from small image patch (tokens in encoder sequence) or target at one specific object instance (tokens in decoder sequence), thus they inherently have the tendency to focus more on local content of the image (with more weight on themselves and their close neighbors during the attention process). 
% As a result, they can not effectively model the global image or inter-object relationships. 
By contrast, domain queries do not need to focus on local features or instances, thus can better aggregate global context and close domain gaps related to scene layout and inter-object relationships without bias.

% \vspace{-5mm}
\subsection{Hierarchical Sequence Feature Alignment}
\label{subsec:proda}

To achieve a more comprehensive feature alignment, we adopt hierarchical feature alignment to progressively align the source and target sequence features in a shallow to deep manner. The hierarchical feature alignment on sequences in the transformer encoder is described as:
\begin{equation}
    \mathcal{L}_{enc}=\sum_{l=1}^{L_{enc}} \left(  \mathcal{L}_{enc_{k}}^{\ell} + \lambda_{enc_{q}} \mathcal{L}_{e n c_{q}}^{\ell} \right),
\end{equation}
where $\lambda_{enc_{q}}$ is a hyperparameter to balance the query-based alignment loss and token-based alignment loss. It is set as 0.1 in our experiments. Similarly, the hierarchical feature alignment is applied on the sequence features in the transformer decoder, $i.e.$,
\begin{equation}
    \mathcal{L}_{dec}=\sum_{l=1}^{L_{dec}} \left(\mathcal{L}_{dec_{k}}^{\ell} + \lambda_{dec_{q}} \mathcal{L}_{dec_{q}}^{\ell} \right),
\end{equation}
where $\lambda_{dec_{q}}$ is a hyperparameter similar to $\lambda_{enc_{q}}$, and is also set as 0.1 in our experiments. For both encoder and decoder, 3-layer MLPs are adopted as the discriminators. Hierarchical feature alignment facilitates better alignment on the sequence features.

\subsection{Bipartite Matching Consistency}
\label{subsec:consist}
Detection transformers~\cite{detr, zhu2020deformable} adopt deep supervision~\cite{lee2015deeply} for training. Auxiliary output are generated on each decoder layer. The auxiliary output on the $\ell$-th decoder layer is denoted as $\hat{y}_{\ell}$, which contains predictions for $M$ object instances. Prediction for each instance includes a class probability vector $\hat{c}_{\ell}^{i}$ and a bounding box prediction $\hat{b}_{\ell}^{i}$. The auxiliary output can be written as:
\begin{equation}
    \hat{y}_{\ell}=\left[\left(\hat{c}_{\ell}^{1}, \hat{b}_{\ell}^{1}\right),\left(\hat{c}_{\ell}^{2}, \hat{b}_{\ell}^{2}\right), \ldots,\left(\hat{c}_{\ell}^{M}, \hat{b}_{\ell}^{M}\right)\right].
\end{equation}

Detection transformers view object detection as a set prediction problem, and use bipartite matching to make one-to-one correspondences between the model output and the ground-truth objects or background class $\varnothing$ in the image~\cite{detr}. Since no semantic label is available on the target domain, the object detector is prone to produce inaccurate matches between object queries and ground-truth objects on the target domain. To tackle this problem, we ensemble the outputs of different decoder layers and constrain the outputs of each decoder layer to produce consistent bipartite matching during training. 
The loss is defined as:
\begin{equation}
    \mathcal{L}_{{cons}}=\frac{1}{L_{{dec}}} \sum_{\ell=1}^{L_{{dec}}} \mathcal{L}_{{cons}}\left(\hat{y}, \hat{y}_{\ell}\right),
\end{equation}
% pseudo 
where $\hat{y}$ is the reference output obtained by averaging the predictions of all decoder layers, $\mathcal{L}_{\text {cons }}$ is the consistency loss that measures the bipartite matching consistency between two predictions. Specifically, it is the combination of JS-divergence between the classification outputs and the L1 distance between bounding box regression outputs, $i.e.$,
\begin{equation}
    \mathcal{L}_{{cons }}\left(\hat{y}_{\ell^{\prime}}, \hat{y}_{\ell}\right)= \frac{1}{M} \sum_{i=1}^{M}\left[\text {JSD}\left(\hat{c}_{\ell^{\prime}}^{i} \| \hat{c}_{\ell}^{i}\right)+\lambda_{\mathrm{L} 1}\left\|\hat{b}_{\ell^{\prime}}^{i}-\hat{b}_{\ell}^{i}\right\|_{1}\right],
\end{equation}
where $\text {JSD}( \cdot \|  \cdot )$ represents JS-divergence, $\lambda_{\mathrm{L} 1}$ is a hyper-parameter to balance the two losses. In this way, we constrain the output of different decoder layers to be consistent and improve the detection transformer's discriminability on the target domain.

% 0person & 3rider & 1car & 4truck & 7bus & 2train & 5mcycle & 6bicycle 
\begin{table*}[t]
% 	\begin{spacing}{1.1}
		\centering
		\small
		\caption{Results of different methods for weather adaptation, $i.e.$, Cityscapes to Foggy Cityscapes. FRCNN and DefDETR are abbreviations for Faster RCNN and Deformable DETR, respectively.} 
		\label{tab:foggy}
		\setlength{\tabcolsep}{3.2mm}
		\begin{tabular}{c|c|cccccccc|c}
			\toprule[1.0pt]
			Method & Detector & person & rider & car & truck & bus & train & mcycle & bicycle & mAP \\
			\hline
			\hline
			Faster RCNN (Source) & FRCNN & 26.9 & 38.2 & 35.6 & 18.3 & 32.4 & 9.6 & 25.8 & 28.6 & 26.9 \\
			DAF \cite{dafaster} & FRCNN & 29.2 & 40.4 & 43.4 & 19.7 & 38.3 & 28.5 & 23.7 & 32.7 & 32.0 \\
			DivMatch \cite{divmatch} & FRCNN & 31.8 & 40.5 & 51.0 & 20.9 & 41.8 & 34.3 & 26.6 & 32.4 & 34.9 \\
			SWDA \cite{strong-weak} & FRCNN & 31.8 & 44.3 & 48.9 & 21.0 & 43.8 & 28.0 & 28.9 & 35.8 & 35.3 \\
			SCDA \cite{scda} & FRCNN & 33.8 & 42.1 & 52.1 & \textbf{26.8} & 42.5 & 26.5 & 29.2 & 34.5 & 35.9 \\
			MTOR \cite{mtor} & FRCNN & 30.6 & 41.4 & 44.0 & 21.9 & 38.6 & 40.6 & 28.3 & 35.6 & 35.1 \\
			CR-DA~\cite{xu2020exploring} & FRCNN & 30.0 & 41.2 & 46.1 & 22.5 & 43.2 & 27.9 & 27.8 & 34.7 & 34.2 \\
			CR-SW~\cite{xu2020exploring} & FRCNN & 34.1 & 44.3 & 53.5 & 24.4 & 44.8 & 38.1 & 26.8 & 34.9 & 37.6 \\
			GPA \cite{gpadet} & FRCNN & 32.9 & 46.7 & 54.1 & 24.7 & 45.7 & \textbf{41.1} & \textbf{32.4} & 38.7 & 39.5 \\
			\hline
			FCOS (Source) & FCOS & 36.9 & 36.3 & 44.1 & 18.6 & 29.3 & 8.4 & 20.3 & 31.9 & 28.2 \\
			EPM~\cite{hsu2020every} & FCOS & 44.2 & 46.6 & 58.5 & 24.8 & 45.2 & 29.1 & 28.6 & 34.6 & 39.0 \\
			\hline
			Deformable DETR (Source) & DefDETR & 37.7 & 39.1 & 44.2 & 17.2 & 26.8 & 5.8 & 21.6 & 35.5 & 28.5 \\
			SFA (Ours) & DefDETR & \textbf{46.5} & \textbf{48.6} & \textbf{62.6} & 25.1 & \textbf{46.2} & 29.4 & 28.3 & \textbf{44.0} & \textbf{41.3} \\
			\bottomrule[1.0pt]
		\end{tabular}
% 	\end{spacing}
	\vspace{-2mm}
\end{table*}

\subsection{Total Loss}
\label{subsec:total}

To summarize, the final training objective of SFA is defined as:
\begin{equation}
\begin{aligned}
  \min_{G} \max_{D}  \mathcal{L}_{{det}}(G) - \lambda_{enc} \mathcal{L}_{enc}(G, D) & - \lambda_{dec} \mathcal{L}_{dec}(G, D) \\& + \lambda_{con} \mathcal{L}_{{cons}}(G),
  \label{eq:overal}
\end{aligned}
\end{equation}
where $G$ is the object detector and $D$ denotes the domain discriminators.
$\lambda_{dec}$, and $\lambda_{cons}$ are hyper-parameters that balance the loss terms. The min-max loss function is implemented by gradient reverse layers~\cite{dann}.Our method is not restricted to specific detection transformer. Instead, it is widely applicable to the family of detection transformers, such as DETR~\cite{detr} and Deformable DETR~\cite{zhu2020deformable}.

\section{Theoretical Analysis}

This section theoretically analyses our method. The performance on the target domain is decomposed into three factors: (1) the expected error on the source domain; (2) the domain divergence between source and target; and (3) the error of the ideal joint hypothesis shared by both domains. 
The domain divergence can further be estimated via a generalization bound.

\subsection{Domain Adaptation Analysis}

According to the theory of domain adaptation by Ben-David \etal~\cite{ben2010theory}, the expected error on the target samples, $R_{\mathcal{T}}(h)$, can be decomposed into three terms, as shown in the following theorem:

\begin{theorem}
\label{th:th_1}
  Let $H$ be the hypothesis class. Given two domains $\mathcal{S}$ and $\mathcal{T}$, we have
    \begin{eqnarray}
    \begin{split}
    %  \begin{gathered}
    \forall h  \in H, R_{\mathcal{T}}(h) &\leq R_{\mathcal{S}}(h)  +\frac{1}{2}{d_{\mathcal{H} \Delta \mathcal{H}}(\mathcal{S},\mathcal{T})}+\lambda. \\
    \end{split}
    \label{eq:decomposition}    
  \end{eqnarray}
Here, $R_{\mathcal{T}}(h)$ is the error of hypothesis $h$ on the target domain, and $R_{\mathcal{S}}(h)$ is the corresponding error on the source domain. ${d_{\mathcal{H} \Delta \mathcal{H}}(\mathcal{S},\mathcal{T})}$ represents the domain divergence that is associated with the feature transferability, and $\lambda =\min _{h \in H} \left[R_{\mathcal{S}}(h)+R_{\mathcal{T}}(h)\right]$ is the error of joint ideal hypothesis that is associate with the feature discriminability. 
\label{th:thm1}
\end{theorem}
In Inequality~\ref{eq:decomposition}, $R_{\mathcal{S}}$ is easily minimized by the supervised loss on source domain. Besides, our token-wise feature alignment minimizes the domain divergence $d_{\mathcal{H} \Delta \mathcal{H}}(\mathcal{S},\mathcal{T})$ and improves the feature transferability. In the meantime, domain-query feature alignment adaptively selects and aligns token features with significant domain gaps while maintaining discriminability for features with smaller domain shifts. Bipartite matching consistency loss ensembles predictions made by multiple decoder layers to obtain a more accurate result. These designs ensure the feature discriminability on the target domain and minimize $\lambda$.

\subsection{Generalization Analysis}

Adversarial training is employed to help learn the mapping from the target domain to the source domain. The generalizability determines the performance of the mapping \cite{mohri2012foundations, he2020recent}.

Denote the generating distributions of the existing data $\mu$ and the generating distribution of the generated data as $\nu$. Denote the empirical counterparts of $\mu$ and $\nu$ are $\hat \mu_{N}$ and $\nu_{N}$, where $N$ is the size of the training sample set. Suppose adversarial training is learning a generator $g \in \mathcal G$ and a discriminator $f \in \mathcal F$, where $\mathcal G$ and $\mathcal F$ are both the hypothesis classes. As described in Section~\ref{subsec:proda}, the discriminator is a three-layer MLP which is constituted by three fully connected layers and three nonlinear operations (nonlinearities),  $(A_{1}, \sigma_{1}, A_{2}, \sigma_{2}, A_{3}, \sigma_{3})$, where $A_{i}$ is a fully connected layer, and $\sigma_{i}$ is a nonlinearity (ReLU). Then we have the following theory.

\begin{theorem}[Covering bound for the discriminator]
	\label{thm:cover_bound}
	Suppose the spectral norm of each weight matrix is bounded: $\|A_{i}\|_{\sigma} \le s_{i}$. Also, suppose each weight matrix $A_{i}$ has a reference matrix $M_{i}$, which is satisfied that $\| A_{i} - M_{i} \|_{\sigma} \le b_{i}$, $i = 1, \ldots, 3$. Then, the $\varepsilon$-covering number satisfies that
	\begin{align}
	\label{eq:cover_bound}
	& \log \mathcal N \left(\mathcal F|_{S}, \varepsilon, \| \cdot \|_{2} \right) \nonumber\\
	\le & \frac{\log\left( 2W^{2} \right) \| X \|_{2}^{2}}{\varepsilon^{2}} \left( \prod_{i = 1}^{3} s_{i} \right)^{2} \sum_{i = 1}^{3} \frac{b_{i}^{2}}{s_{i}^{2}},
	\end{align}
	where $W$ is the largest dimension of the feature maps throughout the algorithm.
\end{theorem}

This theorem is based on \cite{bartlett2017spectrally, he2020why}. A detailed proof is given in the Supplementary Material. Also, Zhang \etal~\cite{zhang2018discrimination} suggests that the generalizability of a GAN~\cite{gan, dann} is determined by the hypothesis complexity of the discriminator. Following this insight, we employ simple discriminators to enhance the generalizability and further enhance the domain adaptation performance.

\begin{table}[t]
% 	\begin{spacing}{1.1}
		\centering
		\small
		\caption{Results of different methods for synthetic to real adaptation, $i.e.$, Sim10k to Cityscapes.} 
		\label{tab:sim10k}
		\setlength{\tabcolsep}{4mm}
		\begin{tabular}{c|c|c} 
			\toprule[1.0pt]
			Methods & Detector & \emph{car} AP \\
			\hline
			\hline
			Faster RCNN (Source) & FRCNN & 34.6 \\
			DAF \cite{dafaster} & FRCNN & 41.9 \\
			DivMatch \cite{divmatch} & FRCNN & 43.9 \\
			SWDA \cite{strong-weak} & FRCNN & 44.6 \\
			SCDA \cite{scda} & FRCNN & 45.1 \\
			MTOR \cite{mtor} & FRCNN & 46.6 \\
			CR-DA~\cite{xu2020exploring} & FRCNN & 43.1 \\
			CR-SW~\cite{xu2020exploring} & FRCNN & 46.2 \\
			GPA  \cite{gpadet} & FRCNN & 47.6 \\
			\hline
			FCOS (Source) & FCOS & 42.5 \\
			EPM~\cite{hsu2020every} & FCOS & 47.3 \\
			\hline
			Deformable DETR (Source) & DefDETR & 47.4  \\
			SFA (Ours) & DefDETR & \textbf{52.6} \\
			\bottomrule[1.0pt]
		\end{tabular}
% 	\end{spacing}
	\vspace{-1mm}
\end{table}

\section{Experimental Results}
\label{sec:experiments}

In this section, we evaluate the proposed SFA on three challenging domain adaptation scenarios. Ablation studies are performed to investigate the impact of each component in SFA. Finally, visualization and analysis are presented for better understanding.

\begin{table*}[t]
% 	\begin{spacing}{1.1}
		\centering
		\small
		\caption{Results of different methods for scene adaptation, $i.e.$, Cityscapes to BDD100k daytime subset.}
		\label{tab:bdd}
		\setlength{\tabcolsep}{3.2mm}
		\begin{tabular}{c|c|ccccccc|c}
			\toprule[1.0pt]
			Methods & Detector & person & rider & car & truck & bus & mcycle & bicycle & mAP \\
			\hline
			\hline
            Faster R-CNN (Source) & FRCNN & 28.8 & 25.4 & 44.1 & 17.9 & 16.1 & 13.9 & 22.4 & 24.1 \\
            DAF~\cite{dafaster} & FRCNN & 28.9 & 27.4 & 44.2 & 19.1 & 18.0 & 14.2 & 22.4 & 24.9 \\
            SWDA~\cite{strong-weak} & FRCNN & 29.5 & 29.9 & 44.8 & 20.2 & 20.7 & 15.2 & 23.1 & 26.2 \\
            SCDA~\cite{scda} & FRCNN & 29.3 & 29.2 & 44.4 & 20.3 & 19.6 & 14.8 & 23.2 & 25.8 \\
            CR-DA~\cite{xu2020exploring} & FRCNN & 30.8 & 29.0 & 44.8 & 20.5 & 19.8 & 14.1 & 22.8 & 26.0 \\
            CR-SW~\cite{xu2020exploring} & FRCNN & 32.8 & 29.3 & 45.8 & 22.7 & 20.6 & 14.9 & 25.5 & 27.4 \\
            \hline
			FCOS~\cite{tian2019fcos} (Source) & FCOS & 38.6 & 24.8 & 54.5 & 17.2 & 16.3 & 15.0 & 18.3 & 26.4 \\
			EPM~\cite{hsu2020every} & FCOS & 39.6 & 26.8 & 55.8 & 18.8 & 19.1 & 14.5 & 20.1 & 27.8 \\
            \hline
            Deformable DETR (Source) & DefDETR & 38.9 & 26.7 & 55.2 & 15.7 & 19.7 & 10.8 & 16.2 & 26.2 \\
            SFA (Ours) & DefDETR &  \textbf{40.2} & 27.6 & \textbf{57.5} & 19.1 & \textbf{23.4} & \textbf{15.4} & 19.2 & \textbf{28.9} \\
			\bottomrule[1.0pt]
		\end{tabular}
% 	\end{spacing}
	\vspace{-2mm}
\end{table*}

\begin{table*}[t]
		\centering
		\small
		\caption{Ablation studies on the Cityscapes to Foggy Cityscapes scenario. CNN represents feature alignment on the CNN backbone. DQ, TW, BMC, and HR represent domain query-based feature alignment, token-wise feature alignment, bipartite matching consistency, and hierarchical feature alignment, respectively. } \label{tab:ablation}
		\setlength{\tabcolsep}{2mm}
		\begin{tabular}{c|ccccc|cccccccc|c}
			\toprule[1.0pt]
			Methods & CNN & DQ & TW & BMC & HR & person & rider & car & truck & bus & train & mcycle & bicycle & mAP \\
			\hline
			\hline
			Deformable DETR (Source) &  &  &  &  &  & 37.7 & 39.1 & 44.2 & 17.2 & 26.8 & 5.8 & 21.6 & 35.5 & 28.5 \\
			\hline
			\multirow{7}{*}{Proposed} & $\checkmark$ &  &  &  &  & 43.8 & 45.6 & 55.5 & 18.2 & 38.7 & 8.4 & 28.1 & 43.3 & 35.2 \\
			&  & $\checkmark$ &  &  &  & 45.1 & 46.8 & 61.3 & 21.6 & 36.6 & 9.5 & \textbf{30.6} & 40.4 & 36.5 \\
			&  &  & $\checkmark$ &  &  & 45.6 & 47.6 & 60.7 & 23.6 & 41.3 & 15.5 & 26.2 & 41.1 & 37.7 \\
			&  &  &  & $\checkmark$ &  & 38.0 & 39.3 & 45.6 & 16.5 & 28.4 & 6.0 & 24.5 & 37.5 & 29.5 \\ 
			&  & $\checkmark$ & $\checkmark$ &  &  & 45.7 & 47.6 & 61.6 & 23.7 & 43.8 & 16.4 & 28.2 & 43.3 & 38.8 \\
			&  & $\checkmark$ & $\checkmark$ &  & $\checkmark$ & 46.0 & 46.7 & 62.4 & 24.1 & 45.6 & 22.1 & 27.6 & 43.9 & 39.8 \\
			& $\checkmark$ & $\checkmark$ & $\checkmark$ &  & $\checkmark$ & 46.3 & 48.2 & 62.2 & 22.1 & 43.4 & 24.3 & 29.9 & 43.1 & 39.9 \\ 
            \hline
			SFA (Ours) &  & $\checkmark$ & $\checkmark$ & $\checkmark$ & $\checkmark$ & \textbf{46.5} & \textbf{48.6} & \textbf{62.6} & \textbf{25.1} & \textbf{46.2} & \textbf{29.4} & 28.3 & \textbf{44.0} & \textbf{41.3} \\
			\bottomrule[1.0pt]
		\end{tabular}
% 	\end{spacing}
	\vspace{-2mm}
\end{table*}

\subsection{Experimental Setup}
\subsubsection{Datasets}
\label{ssec:datasets}
Four public datasets are utilized in our experiments, including Cityscapes~\cite{cityscapes}, Foggy Cityscapes~\cite{foggy}, Sim10k~\cite{sim10k}, and BDD100k~\cite{yu2018bdd100k}, which are detailed as follows.

\begin{itemize}
\setlength{\itemsep}{1pt}
\setlength{\parsep}{0pt}
\setlength{\parskip}{0pt}
	\item \textbf{Cityscapes}~\cite{cityscapes} is collected from urban scenes, which contains 3,475 images with pixel-level annotation. Among them, 2,975 and 500 images are used for training and evaluation, respectively. Bounding box annotation of 8 different object categories can be obtained by taking the tightest rectangles of object masks. 
	\item \textbf{Foggy Cityscapes}~\cite{foggy} is obtained by applying the fog synthesis algorithm to Cityscapes, based on depth maps in the Cityscapes dataset. It inherited the annotations in Cityscapes and is suitable for the evaluation of weather adaptation.
	\item \textbf{Sim10k}~\cite{sim10k} is generated by the Grand Theft Auto game engine, which contains 10,000 training images with 58,701 bounding box annotations for cars. It is suitable for the evaluation of synthetic to real adaptation. 
	\item \textbf{BDD100k}~\cite{yu2018bdd100k} contains 100k images, including 70k training images and 10k validation images annotated with bounding boxes. Following~\cite{xu2020exploring}, we extract the \emph{daytime} subset of BDD100k for the evaluation of scene adaptation. The subset includes 36,728 training images and 5,258 validation images.

\end{itemize}

Based on these datasets, we evaluate the proposed SFA on three domain adaptation scenarios: (1) Weather adaptation, $i.e.$, Cityscapes to Foggy Cityscapes, where the models are trained on cityscape and tested on foggy cityscape; (2) Synthetic to real adaptation, $i.e.$, Sim10k to Cityscapes, where the models are trained on Sim10k and tested on Cityscapes; and (3) Scene Adaptation, $i.e.$, Cityscapes to the daytime subset of BDD100k, where the models are trained on Cityscapes and tested on BDD100k daytime subset. Mean Average Precision (mAP) with a threshold of 0.5 is adopted as the evaluation metric, following~\cite{dafaster}.

\subsubsection{Implementation Details}
\label{ssec:Implementation}
By default, our method is built on Deformable DETR~\cite{zhu2020deformable}. ImageNet~\cite{deng2009imagenet} pre-trained ResNet-50~\cite{resnet} is adopted as the backbone in all experiments. Following Deformable DETR, we train the network using the Adam optimizer~\cite{kingma2014adam} for 50 epochs. The learning rate is initialized as $2 \times 10^{-4}$ and decayed by 0.1 after 40 epochs. The batch size is set as 4 in all experiments. 
Both $\lambda_{enc}$ and $\lambda_{dec}$ are set as 1 for weather adaptation, and 0.01 for other domain adaptation scenarios. Similarly, $\lambda_{cons}$ is set as 1 for weather adaptation, and 0.1 for other scenarios. All our experiments are performed on NVIDIA Tesla V100 GPUs. To show the generality of our method, we also provide results of our method built on DETR~\cite{detr}, as shown in the Supplementary Material.

\subsection{Comparisons with SOTA Methods}
\label{ssec:comparisons}

\subsubsection{Weather Adaptation.}

The object detectors are often required to be applicable under various weather conditions. We use the Cityscape to Foggy Cityscape scenario to evaluate the model's robustness to weather variations~\cite{zhang2019famed}. As shown in Table~\ref{tab:foggy}, SFA significantly improves Deformable DETR's cross-domain performance, achieving a 12.8 absolute gain of mAP50. Moreover, it outperforms all previous domain adaptive object detection methods.

\subsubsection{Synthetic to Real Adaptation.}
% Image and semantic label pairs can be automatically generated by game engines~\cite{sim10k}. 
Training an object detector on synthetic images that can generalize to real-world images is fascinating, as it significantly reduces the labor of data collection and annotation. To this end, we evaluate our SFA on the synthetic to real adaptation scenario, as shown in Table~\ref{tab:sim10k}. It can be seen that SFA improves the source-only Deformable DETR with an over 10\% relative performance gain and outperforms all existing domain adaptive object detection methods. 

\subsubsection{Scene Adaptation.}
In real-world applications like autonomous driving, scene layouts frequently change~\cite{lan2020global}.
The performance of our SFA for scene adaptation is shown in Table~\ref{tab:bdd}. Following~\cite{xu2020exploring}, results on 7 common categories are reported. We can see that SFA outperforms the previous state-of-the-art. Moreover, a systematic improvement can be observed, where SFA improves Deformable DETR on all 7 categories over the source only model.

\begin{figure*}[t]
  \centerline{\includegraphics[width=\linewidth]{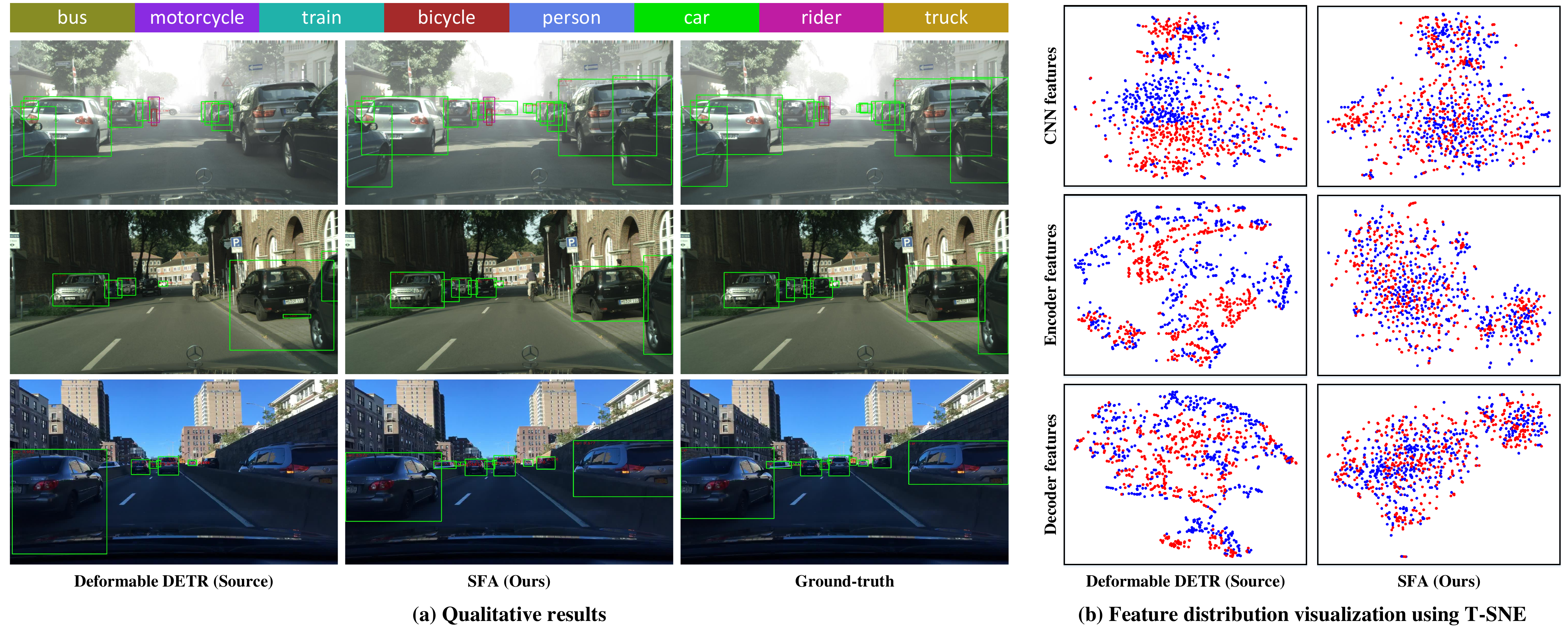}}
   \caption{Visualizations. (a) Qualitative results: From top to bottom are results on the scenarios of Cityscapes to Foggy Cityscapes, Sim10k to Cityscapes, and Cityscapes to BDD100k, respectively. (b) Visualization of feature distributions using T-SNE~\cite{tsne}. The {\color{blue}blue} circles denote the source features, while {\color{red}red} circles represent target features.}
  \label{fig:demo_tsne}
  \Description{activation}
%   \vspace{-3mm}
\end{figure*}

\begin{figure}[t]
  \centerline{\includegraphics[width=\linewidth]{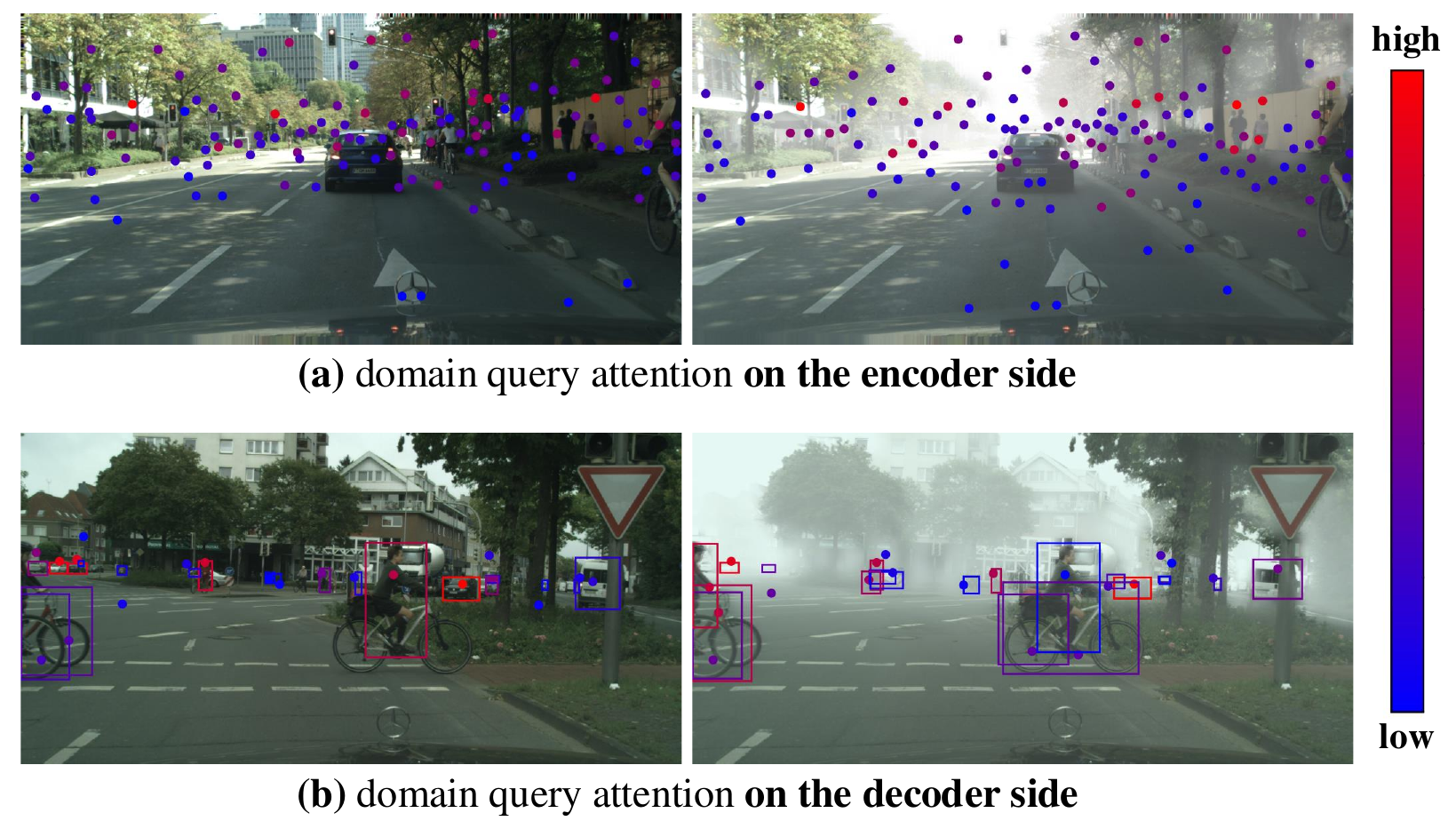}}
  \caption{Visualization of domain queries from both transformer encoder and decoder sides, under the Cityscapes to Foggy Citysccapes scenario. 
  }
  \label{fig:query}
  \Description{activation}
  \vspace{-2mm}
\end{figure}

% \vspace{-2mm}
\subsection{Ablation Studies}
\label{ssec:ablation}
To better understand our method, we conduct ablation studies by isolating each component in SFA, as shown in Table~\ref{tab:ablation}. We have the following observations: (1) both domain query-based feature alignment and token-wise feature alignment can alleviate the domain gaps and improve detection transformer's cross-domain performance by 8.0 and 9.2 mAP, respectively; (2) domain query-based feature alignment and token-wise feature alignment are complementary to each other. Thereby, a combination of both brings further improvement; (3) the hierarchical feature alignment (HFA) is effective and brings a 1.0 mAP gain. Moreover, with HFA, the shallower sequence features near the CNN backbone is aligned. As a result, feature alignment on the CNN backbone can be safely removed without loss of performance; and (4) using the bipartite matching consistency loss alone can bring a 1.0 performance gain. Moreover, it is complementary with the alignment on sequence features, and further improves the model performance by 1.5 mAP.

\subsection{Visualization and Analysis}

\subsubsection{Detection Results}
In Figure~\ref{fig:demo_tsne} (a), we show some visual results by Deformable DETR and our SFA, accompanied with the ground-truth. As can be seen, in all three scenarios, SFA improves the detection performance. It successfully mitigates the false positives generated by Deformanble DETR and detects challenging objects overlooked by Deformanble DETR. Interestingly, from the first row, we can see that SFA successfully detects the distant car that is not labeled in the ground-truth, which further indicates its generalizability to the target domain.

\subsubsection{Visualization of Feature Distribution.}
We present the distribution of features extracted by the CNN backbone, the transformer encoder, and the transformer decoder. As shown in Figure~\ref{fig:demo_tsne} (b), both CNN and sequence features extracted by the source-only Deformable DETR can be easily separated by domain. By contrast, our SFA learns domain-invariant sequence features in both encoder and decoder. Interestingly, the features in the CNN backbone are also aligned in SFA, as the shallower sequence features near the CNN backbone can be aligned by hierarchical feature alignment.

\subsubsection{Visualization of Domain Query.}

The domain query-based feature alignment utilizes a novel domain query to adaptively aggregate and align global context from the sequence feature in the encoder or decoder. It is natural to wonder where the domain query looks at and how much it values each sampling location. To obtain a better understanding of the domain query-based feature alignment, we visualize the sampling locations and attention weights for the domain query. In Figure~\ref{fig:query} (a), it can be seen that the domain query in the encoder learns to attend to sampling locations with obvious domain gaps, \ie, the upper part of the target images with large scene depth dense fog. Moreover, it puts more weight on tokens with more significant domain shifts. A similar phenomenon can be seen on the decoder side, as shown in Figure~\ref{fig:query}(b), the domain query in the decoder put more attention on locations with significant domain gaps. The slight difference is that it focuses more on the foreground objects, \eg, the cars, riders, and bicycles, as the context sequence feature in the decoder models foreground objects. 
% More visualizations are provided in the Supplementary Material.

% \vspace{-4mm}
\section{Conclusion}
\label{sec:conclusion}
In this paper, we focus on making the promising detection transformers domain adaptive. Based on the observation that the feature distribution alignment on CNN backbone does not guarantee domain-invariant features in transformer for prediction, we propose sequence feature alignment (SFA) for detection transformers. Domain query-based feature alignment (DQFA) and token-wise feature alignment (TFA) with explicit technical insights are adopted to close domain gaps at global and local scales, respectively. Moreover, a novel bipartite matching consistency loss is proposed to enhance the feature discriminability for robust object detection. Experimental results validate the effectiveness of our SFA. We hope our approach will serve as a solid baseline and help ease future research on developing domain adaptive detection transformers.

\iffinal

\section*{Acknowledgement}
\label{sec:Acknowledgement}
This work is supported by National Key R\&D Program of China under Grant 2020AAA0105701, National Natural Science Foundation of China (NSFC) under Grants 61872327, 61806062, U19B2038, Major Special Science and Technology Project of Anhui (No. 012223665049), and the University Synergy Innovation Program of Anhui Province under Grants GXXT-2019-025.

\fi

\clearpage

\appendix

% \title{Supplementary Material}
% \maketitle

\section{More Implementation Details}
\subsection{Implementation Details of DA-CNN}
In Figure 1 and Table 4, we presented the results of applying adversarial feature alignment~\cite{dann} on the CNN backbone of Deformable DETR (denoted as DA-CNN in Figure 1). Specifically, we apply hierarchical feature alignment on the CNN backbone, following~\cite{maf}. Feature maps of stages $C_3$ through $C_5$ in ResNet-50~\cite{resnet}, plus one more feature map obtained via a $3 \times 3$ stride $2$ convolution on the final $C_5$ stage are utilized for feature alignment. The sizes of the feature maps are reduced by the information invariant scale reduction modules (SRMs) in~\cite{maf} before seed into the domain discriminators.

We also implemented feature alignment on the CNN backbone following SWDA~\cite{strong-weak}. Specifically, strong local feature alignment on the feature map of stage $C_2$ and weak global feature alignment on the feature map of $C_5$ in ResNet-50 are adopted. We obtain a similar result of 35.5 mAP compared to hierarchical feature alignment in~\cite{maf}. Thus, we only show DA-CNN implemented by hierarchical feature alignment in Figure 1, which is conceptually cleaner.

\subsection{Structure of Domain Discriminators}
As discussed in Section 4.2, simple discriminators can enhance the generalizability and further enhance the domain adaptation performance. Specifically, we simply adopt a 3-layer multilayer perceptron (MLP) structure, as shown in Table~\ref{tab:dis}.
All four discriminators, including $D_{enc_{q}}$, $D_{enc_{k}}$, $D_{dec_{q}}$, and $D_{dec_{k}}$, share the same structure. Besides, $D_{enc_{q}}$ and $D_{enc_{k}}$ share the same weights. Similarly, $D_{dec_{q}}$ and $D_{dec_{k}}$ share the weights. Domain discriminators for encoder and decoder are shared across different feature levels.

\vspace{-3mm}
\begin{table}[h!t]
\caption{The architecture of the domain discriminators.}
\label{tab:dis}
\begin{minipage}[t]{.4\textwidth}
%\footnotesize
\begin{center}
\begin{tabular}{|c|}
\hline
\multicolumn{1}{|c|}{Domain Discriminators} \\
\hline
	Fully Connected $256 \times 256$ \\
	ReLU \\
	Fully Connected $256 \times 256$ \\
	ReLU \\
	Fully Connected $256 \times 2$\\
	Softmax \\
    \hline
\end{tabular}
\end{center}
\end{minipage}
\end{table}

\section{Detailed Theoretical Analysis}

\begin{proof}[Proof of Theorem 4.2]

We denote the spaces of the output functions $F_{(A_{1}, \ldots, A_{i-1})}$ induced by the weight matrices $A_{i}, i = 1, \ldots, 5$ by $\mathcal H_{i}, i = 1, \ldots, 5$, respectively. Lemma A.7 in \cite{bartlett2017spectrally}, suggests inequality,
\begin{align}
	& \log \mathcal N(\mathcal F|S) \nonumber\\
	\le & \log \left( \prod_{i=1}^{5} \sup_{\mathbf A_{i-1} \in \bm {\mathcal B}_{i-1}} \mathcal N_{i} \right) \nonumber\\
	\le & \sum_{i = 1}^{5} \log \left( \sup_{\substack{(A_{1}, \ldots, A_{i-1}) \\ \forall j < i, A_{j} \in B_{j}}} \mathcal N \left( \left\{ A_{i} F_{(A_{1}, \ldots, A_{i-1})} \right\}, \varepsilon_{i}, \| \cdot \|_{2} \right) \right) ~.
\end{align}
We thus get the following inequality,
\begin{align}
	\log \mathcal N(\mathcal F|S) \le \sum_{i = 1}^{5} \frac{b_{i}^{2} \| F_{(A_{1}, \ldots, A_{i-1})}(X) \|^{2}_{\sigma}}{\varepsilon_{i}^{2}} \log \left( 2W^{2} \right) ~. % \nonumber\\
\end{align}
Meanwhile,
\begin{align}
	\| F_{(A_{1}, \ldots, A_{i-1})}(X) \|^{2}_{\sigma} = & \| \sigma_{i-1} (A_{i-1} F_{(A_{1}, \ldots, A_{i-2})}(X)) - \sigma_{i-1}(0) \|_2 \nonumber\\
	\le & \| \sigma_{i-1} \| \| A_{i-1} F_{(A_{1}, \ldots, A_{i-2})}(X) - 0 \|_2 \nonumber\\
	\le & \rho_{i-1} \| A_{i-1} \|_{\sigma} \| F_{(A_{1}, \ldots, A_{i-2})}(X) \|_2 \nonumber\\
	\le & \rho_{i-1} s_{i-1} \| F_{(A_{1}, \ldots, A_{i-2})}(X) \|_2.
\end{align}
Therefore,
\begin{equation}
	\| F_{(A_{1}, \ldots, A_{i-1})}(X) \|^{2}_{\sigma} \le \|X\|^{2} \prod_{j = 1}^{i-1} s_{i}^{2} \rho_{i}^{2}.
\end{equation}

Motivated by the proof given in \cite{bartlett2017spectrally}, we suppose equations:
\begin{gather}
	\varepsilon_{i+1} = \rho_{i} s_{i+1} \varepsilon_{i} ~,\\
	\varepsilon_{5} = \rho_{1} \prod_{i = 2}^{4} s_{i}\rho_{i} s_{5} \epsilon_{1} ~,\\
	\varepsilon = \rho_{1} \prod_{i = 2}^{5} s_{i}\rho_{i} \epsilon_{1} ~.
\end{gather}
Therefore,
\begin{equation}
	\varepsilon_{i} = \frac{\rho_{i}\prod_{j = 1}^{i-1} s_{j}\rho_{j}}{\prod_{j = 1}^{5} s_{j}\rho_{j}} \varepsilon ~.
\end{equation}
Therefore,
\begin{equation}
	\log \mathcal N \left(\mathcal F|_{S}, \varepsilon, \| \cdot \|_{2} \right) \le \frac{\log\left( 2W^{2} \right) \| X \|_{2}^{2}}{\varepsilon^{2}} \left( \prod_{i = 1}^{5} s_{i} \rho_{i} \right)^{2} \sum_{i = 1}^{5} \frac{b_{i}^{2}}{s_{i}^{2}} ~,
\end{equation}
which is exactly Equation 16 of Theorem 4.2.

The proof is completed.
\end{proof}

\begin{table*}[t]
\renewcommand\arraystretch{1.4}
% 	\begin{spacing}{1.1}
		\centering
		\small
		\caption{Sequence feature alignment built on DETR~\cite{detr}. DQ, TW, BMC, and HR represents domain query-based feature alignment, token-wise feature alignment, bipartite matching consistency, and hierarchical feature alignment, respectively. } \label{tab:detr}
		\setlength{\tabcolsep}{3mm}
		\begin{tabular}{c|cccc|cccccccc|c}
			\toprule[1.0pt]
			Methods & DQ & TW & BMC & HR & person & rider & car & truck & bus & train & mcycle & bicycle & mAP \\
			\hline
			\hline
			DETR (Source) &  &  &  &  & 19.4 & 16.0 & 35.2 & 5.1 & 7.8 & 3.0 & 10.5 & 15.9 & 14.1 \\
			\hline
			\multirow{5}{*}{Proposed} & $\checkmark$ &  &  &  & 19.9 & 23.5 & 38.8 & 15.7 & 17.0 & 3.8 & 12.9 & 16.4 & 18.5 \\
			&  & $\checkmark$ &  &  & 18.3 & 24.4 & 41.8 & 14.7 & 22.4 & 5.6 & 9.3 & 17.0 & 19.2 \\
			&  &  & $\checkmark$ &  & 20.7 & 16.5 & 35.5 & 7.6 & 9.0 & 5.0 & 10.3 & 16.0 & 15.0 \\
			& $\checkmark$ & $\checkmark$ &  &  & 20.8 & \textbf{24.5} & 42.0 & 13.0 & 20.3 & 12.9 & 12.7 & 17.7 & 20.5 \\
			& $\checkmark$ & $\checkmark$ &  & $\checkmark$ & 21.2 & 23.5 & 44.0 & 17.6 & 25.2 & 12.7 & 13.9 & 18.0 & 22.0 \\
            \hline
			\bottomrule[1.0pt]
		\end{tabular}
% 	\end{spacing}
	\vspace{-2mm}
\end{table*}

\section{DETR-based SFA}
To show the generalizability of our method, we also implement our Sequence Feature alignment (SFA) based on DETR~\cite{detr}. We note three key difference between the implementation of Deformable DETR-based SFA and DETR-based SFA: (1) hierarchical CNN feature representation of 4 different levels are token as input to the transformer in Deformable DETR, while DETR only uses one feature map at the $C_5$ stage in ResNet; (2) the deformable transformer adopts deformable attention mechanism for the self-attention in the encoder and the cross-attention in the decoder, while DETR adopts the attention mechanism in~\cite{vaswani2017attention}. Therefore, for Deformable DETR, the sample position reference ${p}_{\ell}$ in Equation 1 and Equation 2 are learnable sparse sample locations, while for DETR, ${p}_{\ell}$ simply enumerates all possible locations.

\subsection{Implementation Details}
We evaluate our DETR-based SFA on Cityscapes~\cite{cityscapes} to Foggy Cityscapes~\cite{foggy} scenarios. ImageNet~\cite{deng2009imagenet} pre-trained ResNet-50~\cite{resnet} is adopted as the backbone. Following DETR, we train the network using the Adam optimizer~\cite{kingma2014adam} for 300 epochs. The learning rate is initialized as $10^{-4}$ for the transformer and $10^{-5}$ for the CNN backbone. Learning rates are decayed by a factor of 10 after 200 epochs. The batch size is set as 8 in all experiments. Both $\lambda_{enc}$ and $\lambda_{dec}$ are set as 0.01, $\lambda_{cons}$ is set as 0.1 for other scenarios. All our experiments are performed on NVIDIA Tesla V100 GPUs.

\subsection{Results and Analysis}
The detailed results are shown in Table~\ref{tab:detr}. We observe a much worse baseline, we believe this is due to the deficit of transformer components in processing image feature maps. The attention modules are initialized to cast nearly uniform attention weights to all the pixels in the feature maps at initialization. As a result, more training data and training epochs are required for the attention weights to be learned to focus on sparse meaningful locations. Limited by the number of training images on the Cityscapes dataset (2,975) and a small batch size of 8, DETR shows inferior performance. 

However, as shown in Table~\ref{tab:detr}, our SFA still brings a 9.1 improvement on mAP50 (a relative improvement of 64.5\%) compared to the DETR baseline, which verifies the effectiveness of our method. 
Moreover, ablation studies are provided by isolating each component of SFA. From the results in Table~\ref{tab:detr}, we have the following observations: (1) Similar to the results based on Deformable DETR, both domain query-based feature alignment and token-wise feature alignment can alleviate the domain gaps and improve the detection transformer's cross-domain performance by 4.4 and 5.1 mAP, respectively; (2) domain query-based feature alignment and token-wise feature alignment are complementary to each other. Thereby, a combination of both brings further improvement; (3) the hierarchical feature alignment is effective and brings a 1.5 mAP gain; and (4) using the bipartite matching consistency constraint alone can bring a 0.9 performance gain. Moreover, it is complementary with the feature alignment on sequence features, and further improves the model performance by 1.2 mAP. 

To summarize, our SFA is widely applicable to the detection transformer family, \eg, DETR, Deformable DETR, and significantly improves their cross-domain performance.

\section{More ablation studies}
% \subsection{Ablating DQFA and TWFA.}
As describe in Section 3, both domain-query based feature alignment and token-wise feature alignment can be applied on both encoder and decoder sides. To gain a more comprehensive understanding, we conduct detailed ablation studies on the domain query-based feature alignment and the token-wise feature alignment. Both feature alignment modules applied to the last encoder and decoder layers, without hierarchical feature alignment and bipartite matching consistency loss. 

The detailed results are shown in Table~\ref{tab:more_ablation}, we have the following observations: (1) both domain-query based feature alignment and token-wise feature alignment alleviate the domain gaps between source and target domains, either when they are being applied to the encoder or decoder side. Moreover, they are complementary to each other; (2) the token-wise feature alignment generally shows better performance compare to the domain-query based feature alignment. We assume this is because the domain-query only samples sparse locations for global feature alignment, while the token-wise feature alignment enumerates all possible locations, and enjoys a relatively more comprehensive feature alignment; (3) the feature alignment applied on the encoder side generally performs better it applied to the decoder side. We assume this is because the sequence in encoder side contains both foreground and background features, while the sequence on the decoder side concentrates on modeling the foreground objects. As a result, the feature alignment on the decoder side does not align background feature well, will the feature alignment on the encoder side enjoys a more comprehensive feature alignment. 

\begin{table*}[t]
\renewcommand\arraystretch{1.4}
% 	\begin{spacing}{1.1}
		\centering
		\small
		\caption{Ablation studies on domain query-based feature alignment (DQFA) and token-wise feature alignment (TWFA), \textbf{without hierarchical feature alignment or bipartite matching consistency loss}. Experiments are conducted on the Cityscapes to Foggy Cityscapes Scenario. $\text{DQ}_{enc}$ and $\text{DQ}_{dec}$ indicate the DQFA applied to the last encoder and decoder layers, respectively. Similarly, $\text{TW}_{enc}$ and $\text{TW}_{dec}$ indicate the TWFA applied to the last encoder and decoder layers, respectively} 
		\label{tab:more_ablation}
		\setlength{\tabcolsep}{2mm}
		\begin{tabular}{c|cccc|cccccccc|c}
			\toprule[1.0pt]
			Methods & $\text{DQ}_{enc}$ & $\text{DQ}_{dec}$ & $\text{TW}_{enc}$ & $\text{TW}_{dec}$ & person & rider & car & truck & bus & train & mcycle & bicycle & mAP \\
			\hline
			\hline
			Deformable DETR &  &  &  &  & 37.7 & 39.1 & 44.2 & 17.2 & 26.8 & 5.8 & 21.6 & 35.5 & 28.5 \\
			\hline
			\multirow{8}{*}{Proposed} & $\checkmark$ &  &  &  & 44.8 & 45.9 & 56.9 & 18.7 & 37.8 & 8.5 & 23.1 & 40.3 & 34.5 \\
			&  & $\checkmark$ &  &  & 42.8 & 43.5 & 54.5 & 13.4 & 37.9 & 8.2 & 28.6 & 40.5 & 33.7 \\
			&  &  & $\checkmark$ &  & 44.6 & 48.0 & 58.7 & 24.4 & 39.6 & 9.1 & 29.2 & 43.1 & 37.1 \\
			&  &  &  & $\checkmark$ & 44.9 & 48.1 & 60.3 & 16.2 & 38.0 & 10.7 & 29.0 & 40.9  & 36.1 \\
			& $\checkmark$ & $\checkmark$ &  &  & 45.1 & 46.8 & 61.3 & 21.6 & 36.6 & 9.5 & 30.6 & 40.4 & 36.5 \\
			&  &  & $\checkmark$ & $\checkmark$ & 45.6 & 47.6 & 60.7 & 23.6 & 41.3 & 15.5 & 26.2 & 41.1 & 37.7 \\
			& $\checkmark$ &  & $\checkmark$ &  & 45.6 & 48.2 & 61.0 & 23.0 & 39.8 & 16.3 & 30.1 & 41.0 & 38.1 \\
			&  & $\checkmark$ &  & $\checkmark$ & 46.3 & 47.6 & 61.5 & 18.8 & 42.2 & 15.7 & 26.9 & 41.7 & 37.6 \\
			\bottomrule[1.0pt]
		\end{tabular}
% 	\end{spacing}
	\vspace{3mm}
\end{table*}

\section{More qualitative results}

\begin{figure*}[t]
  \centerline{\includegraphics[width=17.4cm]{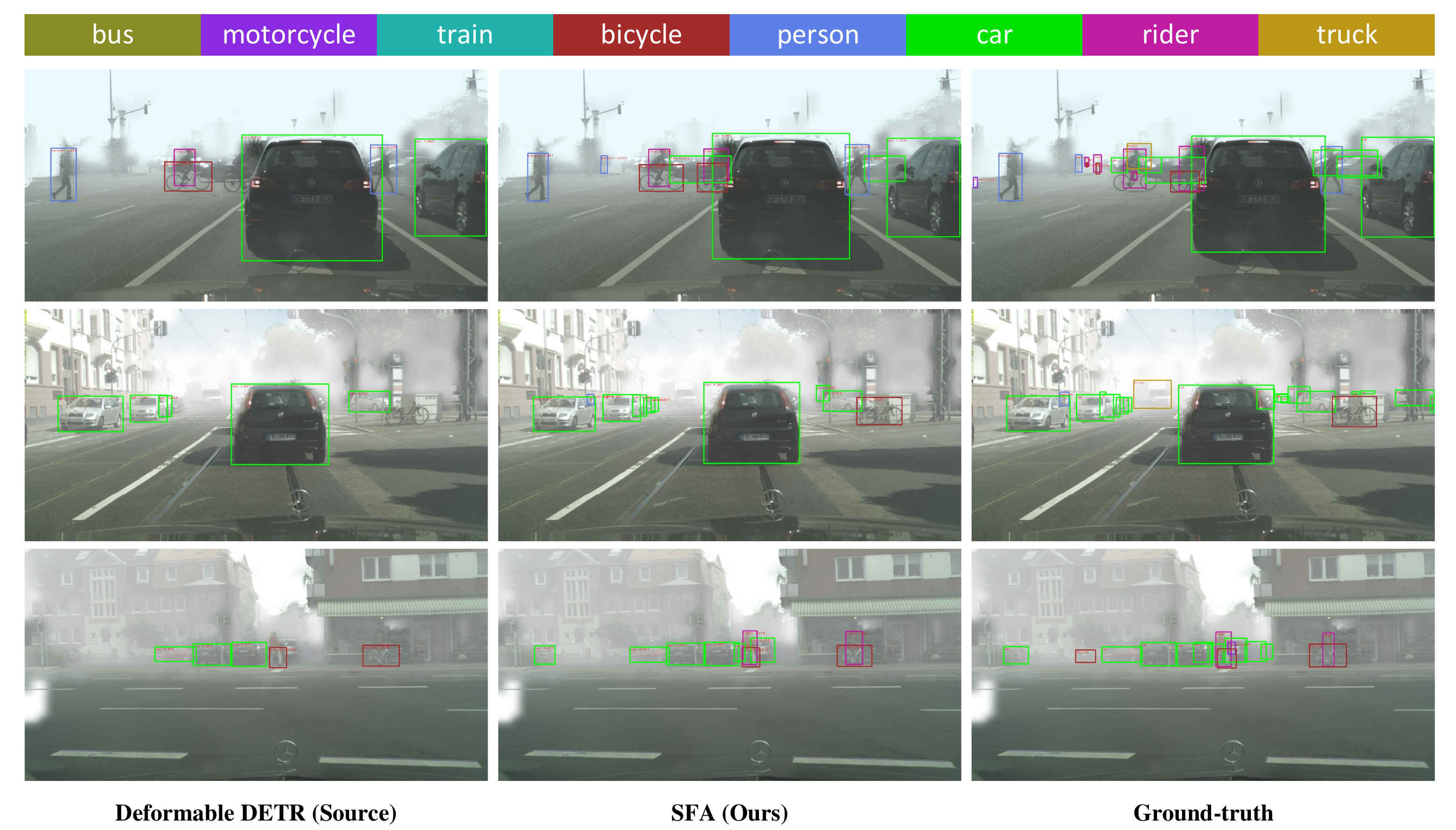}}
   \caption{More qualitative detection results on Cityscapes to Foggy Cityscapes scenario. }
  \label{fig:city}
  \Description{activation}
%   \vspace{-4mm}
\end{figure*}

\begin{figure*}[t]
  \centerline{\includegraphics[width=17.4cm]{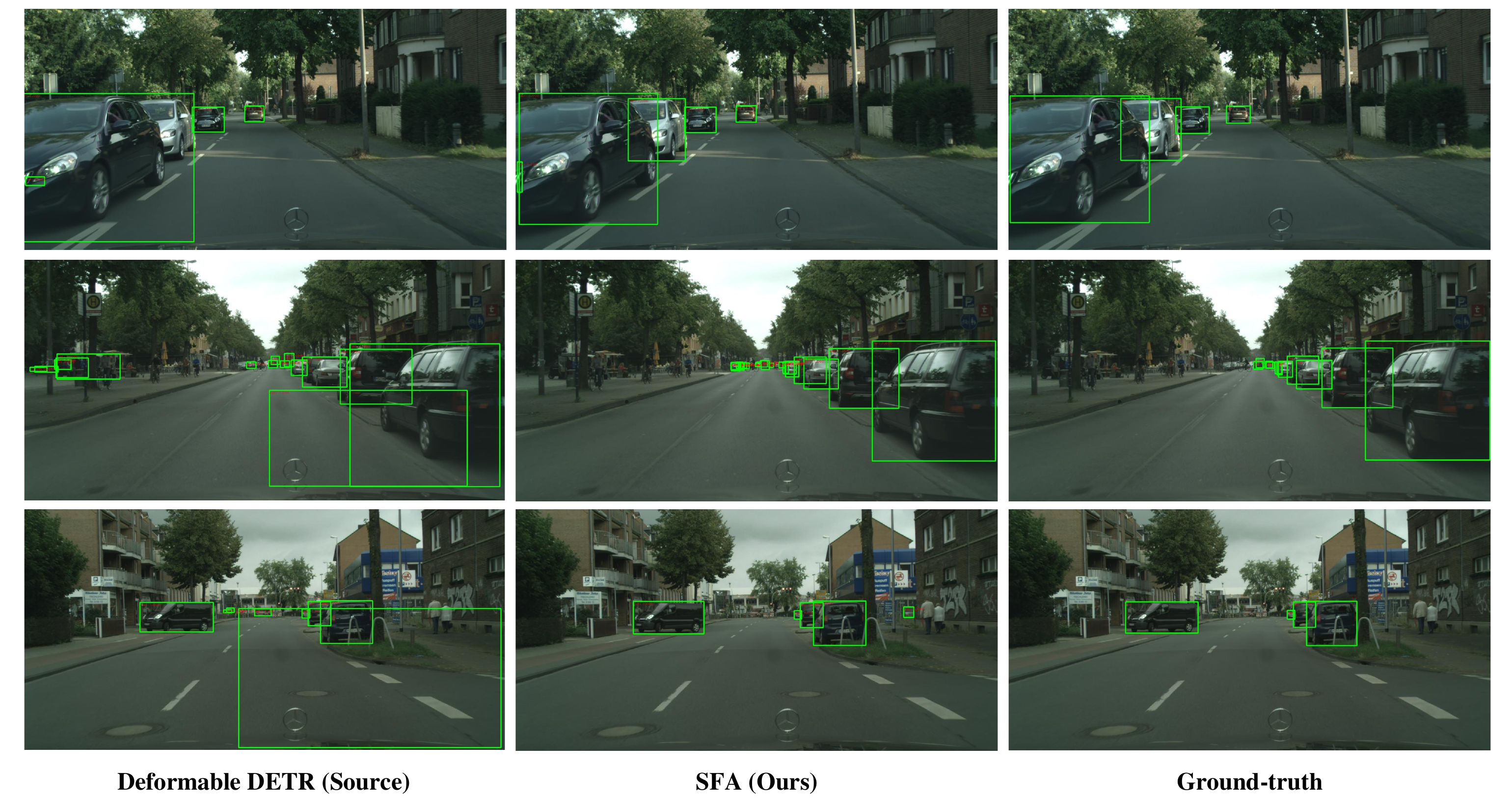}}
  \caption{More qualitative detection results on Sim10k to Cityscapes scenario. }
  \label{fig:sim10k}
  \Description{activation}
%   \vspace{-4mm}
\end{figure*}

\begin{figure*}[t]
  \centerline{\includegraphics[width=17.4cm]{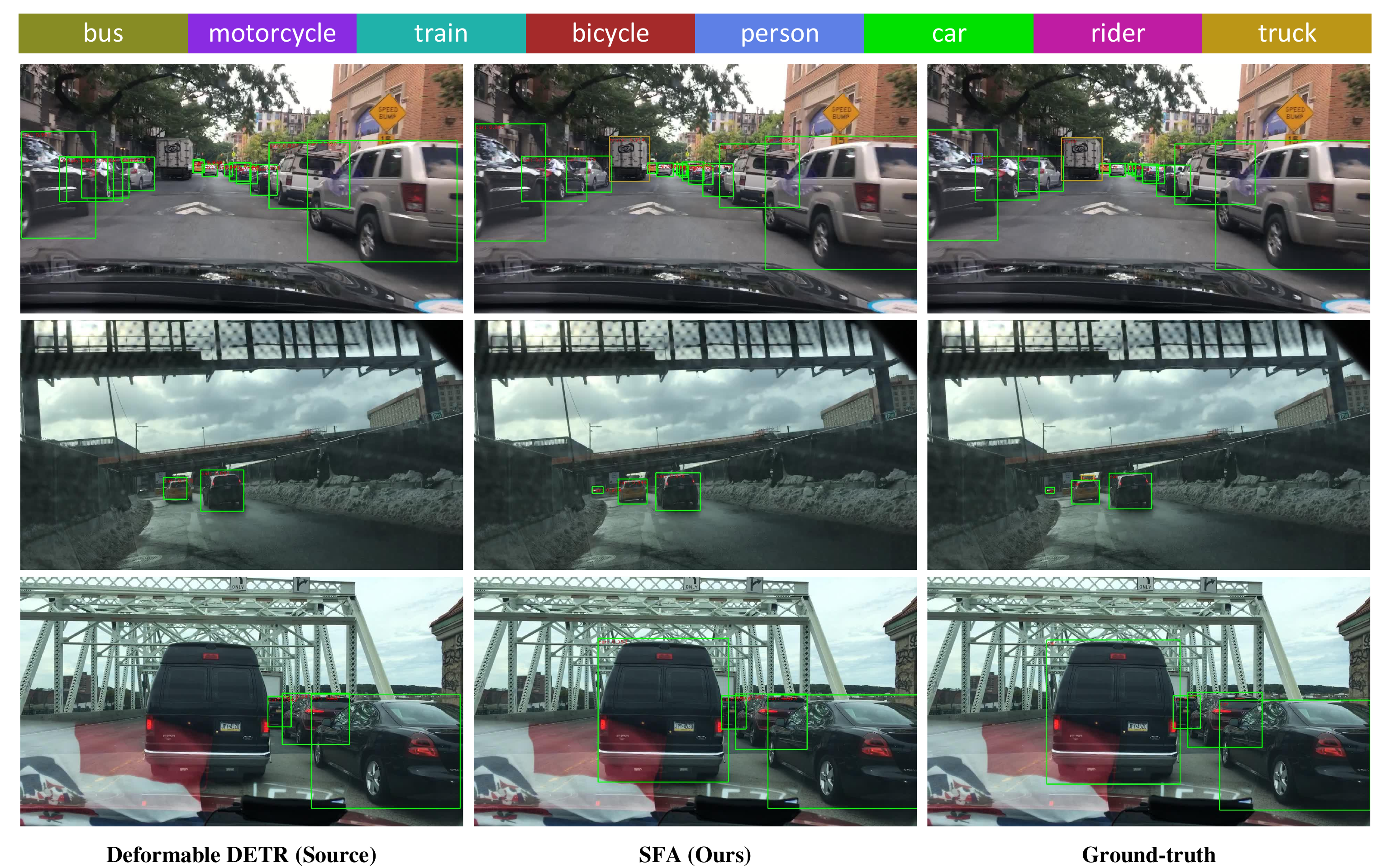}}
   \caption{More qualitative detection results on Cityscapes to BDD100k scenario. }
  \label{fig:bdd}
  \Description{activation}
%   \vspace{-4mm}
\end{figure*}

% \subsection{More Qualitative Results}
More qualitative results sampled from the Cityscapes to Foggy Cityscapes, Sim10k to Cityscapes, and Cityscapes to BDD100k adaptation are shown in Figure~\ref{fig:city},~\ref{fig:sim10k} and~\ref{fig:bdd}, respectively.
Detection results produced by the source-only Deformable DETR and our SFA, accompanied with the corresponding ground-truth are presented. As can be seen, in all three scenarios, SFA improves the detection performance. It successfully mitigates the false positives generated by Deformanble DETR and detects challenging objects overlooked by Deformanble DETR.

% \hspace*{\fill}
% \vspace{4mm}

\clearpage

\bibliographystyle{ACM-Reference-Format}
\balance
\bibliography{conferences,acmart}

\end{document}